\documentclass{article}

\usepackage{microtype}
\usepackage{graphicx}
\usepackage{subfigure}
\usepackage{booktabs} % for professional tables
\usepackage{multirow}

\usepackage{hyperref}

\usepackage[accepted]{icml2023}

\usepackage{amsmath}
\usepackage{amssymb}
\usepackage{mathtools}
\usepackage{amsthm}

\usepackage[capitalize,noabbrev]{cleveref}
\usepackage{multibib}
\newcites{latex}{References}
\theoremstyle{plain}

\theoremstyle{definition}

\theoremstyle{remark}

\usepackage[textsize=tiny]{todonotes}

\icmltitlerunning{Scaling of Class-wise Training Losses for Post-hoc Calibration}

\begin{document}

\twocolumn[
\icmltitle{Scaling of Class-wise Training Losses for Post-hoc Calibration}



\icmlsetsymbol{equal}{*}

\begin{icmlauthorlist}
\icmlauthor{Seungjin Jung}{cauai,comp}
\icmlauthor{Seungmo Seo}{cauai}
\icmlauthor{Yonghyun Jeong}{comp}
\icmlauthor{Jongwon Choi}{cauai,gsaim}
\end{icmlauthorlist}

\icmlaffiliation{cauai}{Department of Artificial Intelligence, Chung-Ang University, Seoul, Korea}
\icmlaffiliation{gsaim}{Department of Advanced Imaging, Chung-Ang University, Seoul, Korea}
\icmlaffiliation{comp}{Naver CLOVA, Seongnam, Korea}

\icmlcorrespondingauthor{Jongwon Choi}{choijw@cau.ac.kr}
\icmlkeywords{Machine Learning, ICML}

\vskip 0.3in
]



\printAffiliationsAndNotice{}  % leave blank if no need to mention equal contribution

\begin{abstract}
The class-wise training losses often diverge as a result of the various levels of intra-class and inter-class appearance variation, and we find that the diverging class-wise training losses cause the uncalibrated prediction with its reliability.
To resolve the issue, we propose a new calibration method to synchronize the class-wise training losses.
We design a new training loss to alleviate the variance of class-wise training losses by using multiple class-wise scaling factors.
Since our framework can compensate the training losses of overfitted classes with those of under-fitted classes, the integrated training loss is preserved, preventing the performance drop even after the model calibration.
Furthermore, our method can be easily employed in the post-hoc calibration methods, allowing us to use the pre-trained model as an initial model and reduce the additional computation for model calibration.
We validate the proposed framework by employing it in the various post-hoc calibration methods, which generally improves calibration performance while preserving accuracy, and discover through the investigation that our approach performs well with unbalanced datasets and untuned hyperparameters.
% We validate the proposed framework by employing it in the various post-hoc calibration methods, which generally improves calibration performance while preserving accuracy.
% We also discover through the investigation that our approach performs well with unbalanced datasets and untuned hyperparameters.
\end{abstract}

\section{Introduction}
With the advancement of deep learning algorithms ~\cite{krizhevsky2017imagenet,yue2020robustscanner,simonyan2014very,girshick2015fast,cheng2020panoptic,li2021dynamic,lin2017focal}, many machine learning tasks have accomplished impressive performance improvement.
However, the safety-critical industries, such as medical and autonomous driving industries, need calibrating prediction beyond high performance to avoid unpredictable errors~\cite{rao2018deep, yao2020comprehensive}.
To tackle the issue, model calibration is widely studied to calibrate the prediction of deep learning models to represent the reliability of the prediction~\cite{TS, muller2019does, mukhoti2020calibrating}.

\begin{figure}
        \centering
        \includegraphics[width=0.95\linewidth]{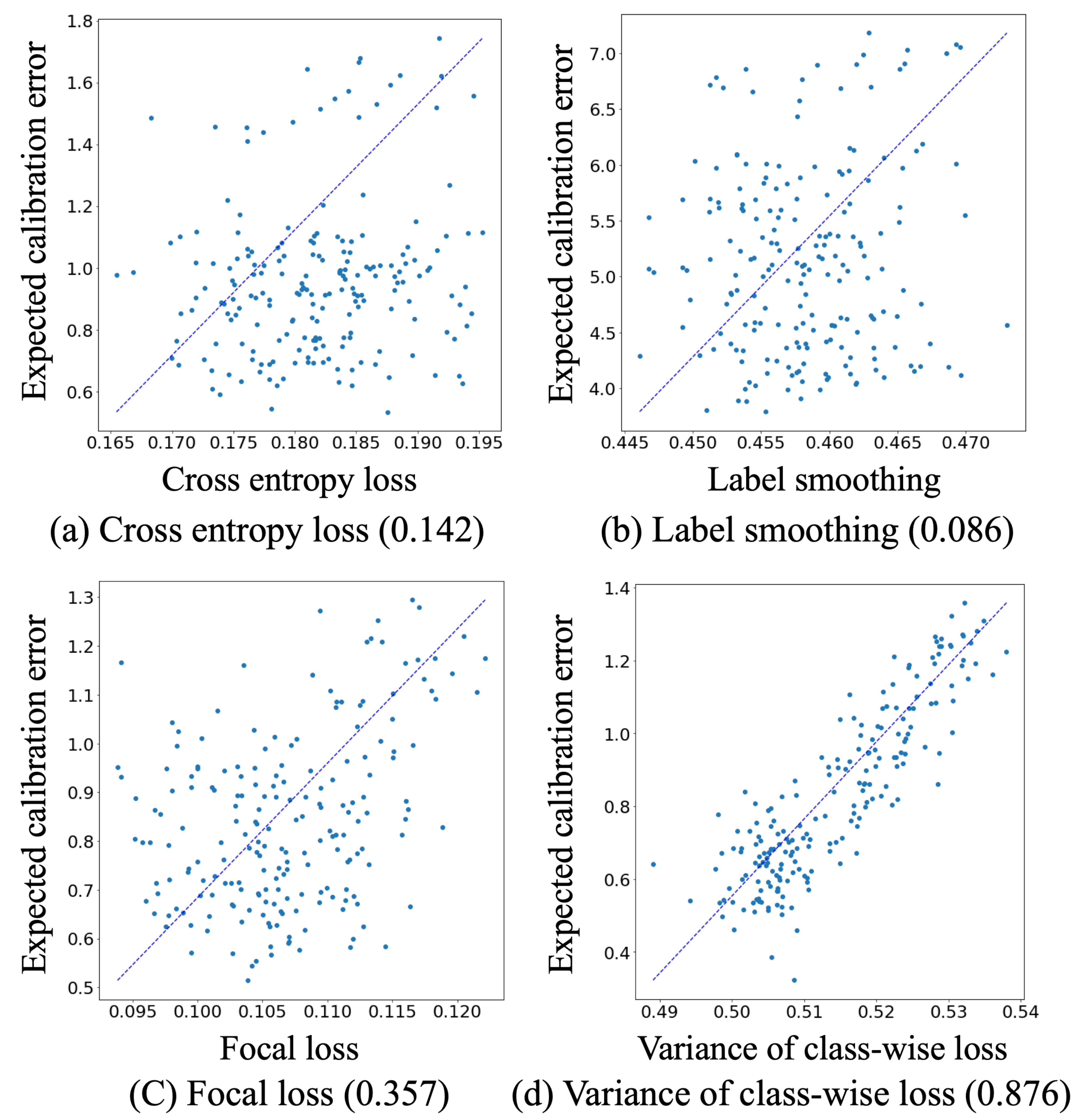}
\vspace{-0.2cm}
\caption{\label{fig:teaser}\textbf{Correlation between calibration error and each loss.} The more linear the correlation coefficient between expected calibration error and loss, the lower the calibration errors. From the plots, we confirm that the variance of class-wise losses shows the highest correlation with the calibration error. Thus, by synchronizing the class-wise losses, we can obtain a well-calibrated prediction from a model. The parentheses in the captions indicate the correlation coefficient.
}
\vspace{-0.2cm}
\end{figure}

However, the existing studies~\cite{TS, muller2019does, mukhoti2020calibrating} have required large additional computations to calibrate the predictions during the training of target models, and further, they suffered from the performance drop to acquire the well-calibrated predictions.
To reduce the additional computation of model calibration, post-hoc calibration studies have also been studied~\cite{PlattCalibration, TS}, which calibrate the prediction by using the weight parameters of pre-trained models.
However, the post-hoc calibration methods suffer from performance drops because the overconfident prediction of a pre-trained model should be recovered back by a limited change of parameters.

To tackle the issue, we investigate that the asynchronous class-wise training losses are critical for preventing target models from being calibrated well.
As partly shown in Fig.~\ref{fig:teaser}, we discover that the variance of class-wise losses is highly correlated to the calibration errors.
The asynchronous class-wise training losses conventionally happen during the training of deep learning models due to the variety of class-wise degrees of intra-class appearance variations and inter-class appearance similarities.
For the classes containing diverse appearance changes, their training loss would be hard to be reduced, and the same situation happens for the classes that are hard to be distinguished from each other.

Based on the discovery, we propose a novel calibration mechanism to synchronize the class-wise training losses.
We design the calibration framework to easily be employed in the post-hoc calibration methods, reducing the additional computation for model calibration by using the pre-trained model as an initial model.
Since the proposed method controls the scale of class-wise training losses, we can apply our method by simply changing the calibration functions of the previous studies to the weighted sum of class-wise training losses.
Additionally, by compensating the training losses of overfitted classes with those of under-fitted classes, our method can maintain the entire training loss, preventing the performance drop after model calibration.

We perform experiments employing our method to various post-hoc calibration methods, which shows the state-of-the-art calibration performance while preserving the accuracy of the original model.
We further validate the real-world applicability of our method by performing experiments with imbalanced datasets.
%We also performed extra experiments to verify the proposed method, considered relationships between class-wise losses, works well on imbalanced datasets.
In addition, we find that the proposed algorithm exhibits stable convergence of calibration performance even when its hyperparameters are fixed during updates, in contrast to previous post-hoc calibration methods requiring schedulers.
The contribution of this work can be summarized as follows:

\begin{itemize}
    \item We verify the effectiveness of the synchronized class-wise losses for improving the prediction reliability while preserving the overall accuracy.
    \item We propose a new post-hoc calibration method that is called class-wise loss scaling to synchronize the class-wise training losses by controlling the loss scale factors.
    \item We validate the proposed method by combining it with various baselines and benchmark datasets, which shows the state-of-the-art calibration performance even while sometimes enhancing the accuracy.    
\end{itemize}

\begin{figure*}[t]
\centering
\includegraphics[width=0.93\linewidth]{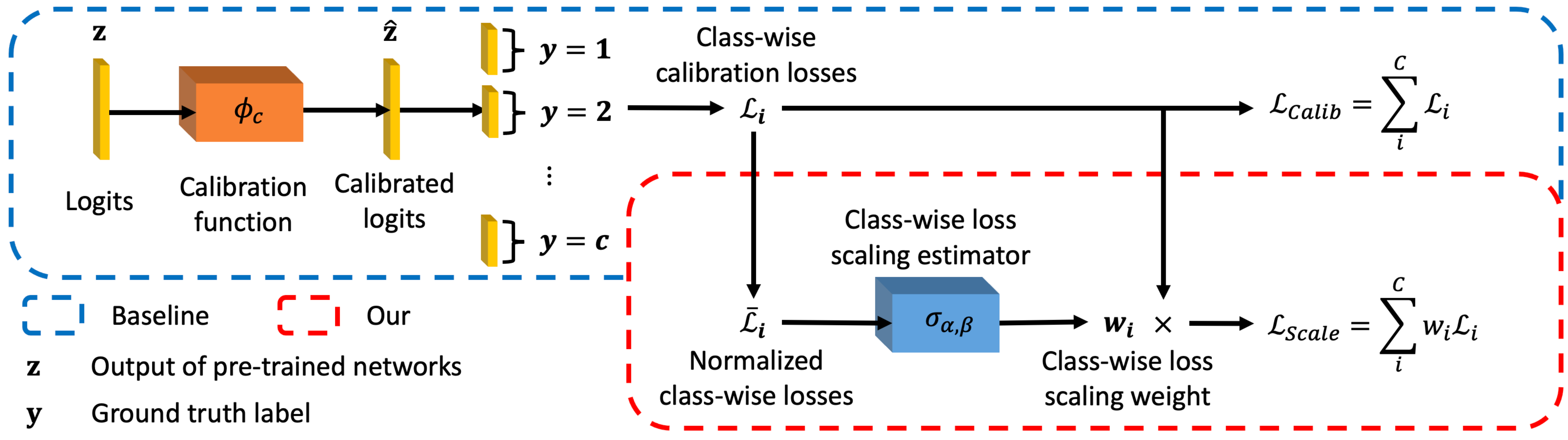}
\caption{\textbf{Overall framework.} Our method brings a class-wise loss and normalizes the class-wise loss to consider relative confidence. Then normalized class-wise loss is used to get class-wise loss scale weight through the estimator. Learning of calibration function is controlled by the estimated class-wise loss scale factor.}
\label{fig:Framework}
\end{figure*}

\section{Related Works}
Typically, Negative Log Likelihood (NLL) is used as training loss in most calibration methods~\cite{TS,Kull2019BeyondTS,rahimi2020intra,ETS,CTS,PTS}. Confidence calibration categorizes into two parts: model calibration and post-hoc calibration.

\subsection{Model Calibration}
%Typically most calibration methods use Negative Log Likelihood (NLL)~\cite{TS} loss to calibrate the logit on the validation set.
Model calibration calibrates the model's confidence during training by preventing under and over confidence.
Guo et al.~\cite{TS} showed that transformations such as model capacities, batch normalization, and weight decay affect the model confidence.
% Guo et al.~\cite{TS} showed how to improve calibration error by altering network depth, length, and weight decay.}
Label Smoothing (LS)~\cite{muller2019does} smooths a one-hot label by decreasing the value of the ground truth label and increasing the values of other class labels. LS achieves model calibration by preventing overfitting on datasets.
% However, Müller et al.~\cite{muller2019does} thought one hot label caused overconfidence, so he proposed the usage of label Smoothing, usually used to model generalization to improve the calibration of prediction probability.
Recently, Mukhoti et al.~\cite{mukhoti2020calibrating} proposed the usage of Focal Loss (FL), which is used for imbalanced datasets. 
FL makes loss magnitude low to the easy datasets and high to the hard datasets because the easy dataset is easy to over-confident and vice versa. 

However, model calibration suffers from a trade-off between confidence and accuracy. Consequently, the increased confidence of the calibrated model comes at the expense of reduced accuracy when compared to a model specifically trained for improved accuracy.

\subsection{Post-hoc Calibration}
%Post-hoc calibration reuses the pre-trained neural networks to calibrate prediction probability by rescaling the output from pre-trained ones.
Post-hoc calibration adjusts the output logits of a pre-trained model that was trained without any consideration for the calibration method.
As a typical traditional calibration method, Platt scaling~\cite{PlattCalibration} is the parametric approach to transform the logits. 
Temperature Scaling (TS)~\cite{TS} and Ensemble Temperature Scaling (ETS)~\cite{ETS} are parametric approaches to calibrate the low-confident logits while preserving accuracy. 
TS only uses the single scalar parameter to generalize data distribution. 
ETS uses more parameters to increase the calibration function's expressive by using ensemble models.
Parameterized Temperature Scaling (PTS)~\cite{PTS} recently has been proposed to estimate appropriate temperature parameter T through the multi-neural layers.
Even though some cases lose the pre-trained networks' performance, Class-based Temperature Scaling (CTS)~\cite{CTS} shows the probability of improving accuracy and calibration.

While temperature-based methods are simple and easy to keep accuracy preserving, matrix-based methods~\cite{Kull2019BeyondTS,rahimi2020intra} are vice versa.
Dirichlet with Off-Diagonal and Intercept Regularisation (DirODIR)~\cite{Kull2019BeyondTS} proposed a matrix with regularization.
Intra Order invariant (IO)~\cite{rahimi2020intra} proposed the order-preserving functions, which retain the top-k predictions of any deep network when used as the post-hoc calibration.

Mukhoti et al. demonstrated the empirical effectiveness of post-hoc calibration, even for models trained with calibration methods~\cite{mukhoti2020calibrating}. However, previous methods overlooked the significance of well-balanced confidences across multiple classes by utilizing a single overall confidence value for the entire dataset.
To address this issue, our method introduces a class-wise loss scaling scheme that takes into account the differences among the confidences of multiple classes.

% \begin{figure*}[t]
% \centering
% \includegraphics[width=0.93\linewidth]{Framework.pdf}
% \caption{\textbf{Overall framework.} Our method brings a class-wise loss and normalizes the class-wise loss to consider relative confidence. Then normalized class-wise loss is used to get class-wise loss scale weight through the estimator. Learning of calibration function is controlled by the estimated class-wise loss scale factor.}
% \label{fig:Framework}
% \end{figure*}

\section{Preliminaries}
Typical calibration methods are training the networks to calibrate well by changing depth, width, weight decay, and loss function~\cite{TS,muller2019does,mukhoti2020calibrating}. 
However, these methods need a high cost to train deep neural networks again.
Post-hoc calibration reuses the pre-trained neural networks to calibrate prediction probability by controlling the uncertainty.
This method can train low cost due to resuing the pre-trained networks.

In the Post-hoc problem, the dataset is separated into the training, validation, and test datasets.
Pre-trained networks use the training datasets to learn the model parameters.
Then, we get logits of the validation datasets and test datasets through the pre-trained networks.
Post-hoc calibration method uses validation and tests logits to train and verify, respectively.

%\subsection{Definition of calibration} 
We address the issue of calibrating neural networks as a parametric method.
We define a $d$-dimensional domain space by $\mathcal{X} \subset \mathbb{R}^d$, and $\mathcal{Y}=\{1,...,n\}$ means a label space with $n$ dimensions.
$\Delta_n$ is an $n-1$ dimensional unit simplex, defined as $\{\mathbf{x}\in\mathbb{R}^n:x_0+...+x_{k-1}=1,x_i\geq0, \forall i = 0,...,k-1\}$.
Then, we can derive a dataset $\mathcal{D}$ of $\mathcal{X\times Y}$ where $X\in\mathcal{X}$ is composed of independent and identically distributed samples from an unknown distribution with $Y\in\mathcal{Y}$. 
We assume that the trained $n$-class probability predictor $\phi_o:\mathbb{R}^d\rightarrow\Delta_n$ is given as $\phi_o=\mathbf{sm}\circ\mathbf{g}$, where $\mathbf{sm}$ represents the sigmoid function and $\mathbf{g}$ is $n$-way classifier resulting in an $n$-way vector $z$.
Then, we can get the predictor output as $\hat{\mathbf{p}}=\phi_o(\mathbf{x})$ and its predicted label is $\hat{y}=\arg\max\;\hat{\mathbf{p}}$.
When $\phi_o$ and $\mathcal{D}$ are given, with the notation of the post-hoc calibration function $\mathbf{f}:\mathbb{R}^n\rightarrow\mathbb{R}^n$, we can derive the calibrated $n$-class probability predictor as $\phi_c\equiv \mathbf{sm}\circ\mathbf{f}\circ\mathbf{g}$.

Then, our goal is to train the post-hoc calibration function $\mathbf{f}$ satisfying Definition~1.

\textbf{Definition 1.}\label{def:cal} 
\textit{
For the distribution $\pi$ of $\mathcal{X}\times\mathcal{Y}$, let random variables $\mathtt{x}\in\mathcal{X}$ and $y\in\mathcal{Y}$ be drawn from $\pi$, when we denote an $n$-class probability predictor by $\phi_c: \mathbb{R}^d\rightarrow\Delta_n$ with the definition of $\hat{y}:=\arg\max \hat{\mathbf{p}}$ and $\hat{\mathbf{p}}:=\phi_c(\mathbf{x})$, we call the $\phi_c$ is in the perfectly calibrated state if satisfying the following equation:
\begin{equation}
Prob(\hat{y}=y|\hat{\mathbf{p}}=\mathbf{p})=\mathbf{p}, \,\,\,\,\, for\;\forall p\in [0,1],
\end{equation}
}
where $p$ is a component of $\mathbf{p}$.
To achieve the perfect calibration, many previous studies~\cite{TS, rahimi2020intra, ETS} have utilized the Negative Log Likelihood (NLL) loss, and we also follow them.

\begin{figure*}
        \centering
        \includegraphics[width=\linewidth]{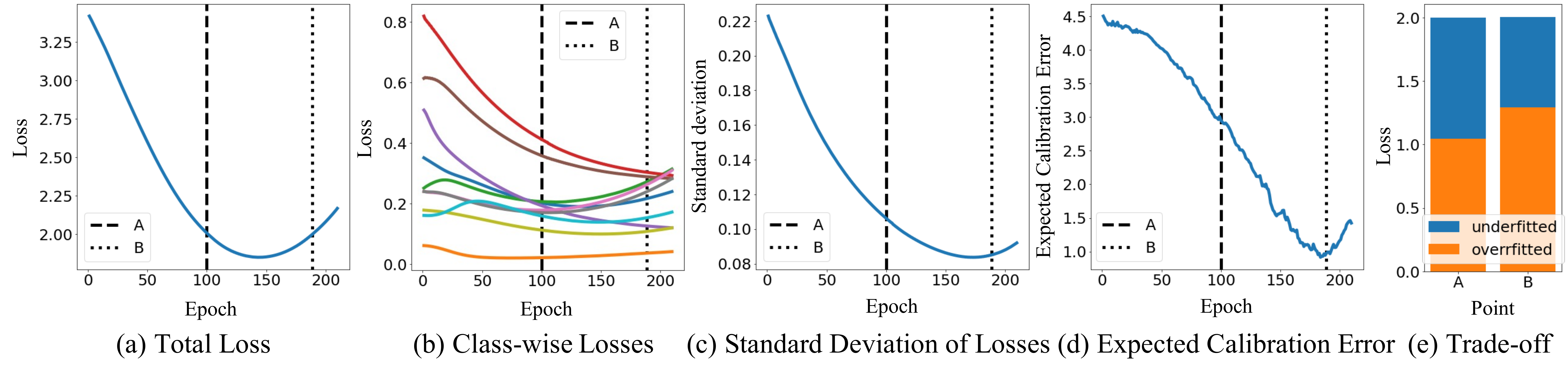}
\caption{\textbf{Analysis of calibration error and class-wise losses.} (a), (b), (c), and (d) show the change of total loss, class-wise losses, a standard deviation of class-wise losses, and expected calibration error by epoch, respectively. A and B represent the points containing the same total loss, but the class-wise losses are diverse in A compared to B. From the results, we can confirm that the expected calibration error can be reduced by minimizing the variance of class-wise losses while preserving the total loss. (e) shows the loss proportion of three under-fitted classes and the other over-fitted classes.}
\label{fig:toy}
\end{figure*}

\section{Class-wise Loss Scaling}
In this section, we describe our new calibration methods for post-hoc calibration, which is named class-wise loss scaling.
The overall framework is described in Fig.~\ref{fig:Framework}.
We first present an analysis of the high correlation between the calibration error and the variance of class-wise training losses.
Then, based on the analysis, we define a class-wise calibration loss, and the class-wise loss scaling estimator is proposed to control the variance of class-wise training losses.

\subsection{Calibration Error and Class-wise Losses}\label{section:4.1}

For the analytical intuition, we focus on the relationship of class-wise training losses instead of the total loss.
The generic strategy to validate the trained model is to the compare prediction of total samples with the corresponding ground-truth data~\cite{foody2017impacts}.
Because this approach only considers the mean accuracy of total samples, the prediction of several classes could be over-confident~\cite{TS, ETS}.
To show the importance of class-wise training losses, we assume that even though two models have the same total training loss, their calibration error would be diverse according to the variance of the class-wise losses.
To examine our strategy, we conducted a simple toy experiment with the post-hoc calibration method using one extra parametric layer.
% Using the pre-trained model, we choose the \textit{top-3} classes with the highest class-wise training losses.
We choose the \textit{top-3} classes with the highest class-wise training losses because we have noticed that these classes exhibit the most distinct trade-off among various top-k classes.
Then, we train the extra parametric layer by using only the class-wise training losses of the chosen classes, which can be denoted as follows:
\begin{equation}
    \mathcal{L} = -\sum_{i\in S}\sum_{(\mathbf{x},y)\in \mathcal{D}_i} \delta(y)^T \log(\phi_c(\mathbf{x})),
\end{equation}
where $S$ and $\mathcal{D}_i$ are the index set of the \textit{top-3} classes and the sample set of $i$-th class, respectively, and $\delta$ transforms the given scalar to the corresponding one-hot vector.
We conduct the toy experiment using the uncalibrated WideResNet-32~\cite{BMVC2016_87} on CIFAR dataset~\cite{Krizhevsky_2009_17719}.

In Fig.\ref{fig:toy}-(a-d), we show the change of the total loss, class-wise loss, standard deviation of losses, and Expected Calibration Error (ECE)~\cite{ECE} according to the training epochs.
We remark on two points, each noted by A and B, which have the same total losses at different training epochs.
We can find that, although the two points have the same total losses, the ECE values are diverse, which verifies that the total training loss is irrelevant to the calibration error.
Thus, the minimization of total loss cannot certainly train the model to be well-calibrated.

In Fig.~\ref{fig:toy}-(b), we can observe the large difference among the class-wise training losses at epoch 0. 
Thus, the variety of class-wise training loss shows that the classes suffer from different types of issues, including over-confident and under-confident predictions.
Meanwhile, the variance of the class-wise training losses becomes much reduced after the additional training epochs, like the ECE values of B.
We also quantify the variance of class-wise training losses by their standard deviation, as shown in Fig.~\ref{fig:toy}-(c).
From the similar trends between Fig.~\ref{fig:toy}-(c) and (d), we can verify that the ECE values are highly correlated to the standard variation of class-wise training losses during the model training.

When we visualize the composition of the total loss at \textit{A} and \textit{B} points as shown in Fig.~\ref{fig:toy}-(e), we can easily find the trade-off between the class-wise training losses of under-fitted and overfitted classes.
The quantity of the reduced class-wise training loss of the under-fitted classes is added to the class-wise training loss of the overfitted classes, which results in the reduced calibration error while preserving the total loss.

From the analysis, we can verify that the variance of the class-wise training loss is a significant factor in controlling the degree of post-hoc calibration.
Furthermore, because the class-wise training losses can be complemented between the overfitted and under-fitted class with the preserved total loss, we can avoid the pre-trained model's performance drop after the post-hoc calibration.
Based on the investigation, we develop a novel mechanism to control the class-wise training losses for post-hoc calibration.

\subsection{Class-wise Calibration Loss}
We use class-wise calibration loss to consider relative confidence between each class.
% The overall framework is described in Fig.~\ref{fig:Framework}.
We define the sample set of class $i$ by $\mathcal{D}_i$, and thus $\mathcal{D}_1 \cup \mathcal{D}_2 ... \cup\mathcal{D}_n =\mathcal{D}$.
We divide the sample set according to their ground truth, so all the samples labeled by the $i$ class are added into $\mathcal{D}_i$.
Then, the class-wise training loss for $i$-th class can be derived as follows:
\begin{equation}
\mathcal{L}_{C_i}= -\sum_{(\mathbf{x}, y)\in\mathcal{D}_i} \delta(y)\log(\phi_c(\mathbf{x})),
\label{eq:7}
\end{equation}
where $\delta$ transforms the given scalar to the corresponding one-hot vector and $\phi_c$ is calibration function, such as TS~\cite{TS}, ETS~\cite{ETS}, PTS~\cite{PTS}, and CTS~\cite{CTS}.
class-wise calibration losses are used to train calibration functions and calculate class-wise confidence levels.

\subsection{Class-wise Loss Scaling Estimator}
In this section, we describe the class-wise loss scale estimator.
Based on the motivation obtained from our analysis, we design the class-wise calibration loss by the weighted sum of the class-wise training losses.
However, we cannot manually assign the class-wise weights because the over-confident and under-confident classes are unknown and differ according to the datasets and network architectures.
To solve the problem, we design a new sigmoid-based function to control the class-wise weights according to their training loss values, which accelerates the trade-off among the class-wise training losses.

First, we normalize the class-wise training losses by
\begin{equation}
\bar{\mathcal{L}}_{C_i}=\frac{\mathcal{L}_{C_i}-\mathbb{E}[\mathcal{L}_{C_i}]}{\sqrt{\mathbb{E}[\mathcal{L}_{C_i}^2]-\mathbb{E}^2[\mathcal{L}_{C_i}]}},
\end{equation}
where the losses are normalized by a normal distribution of zero mean and unit standard deviation.

Then, from the normalized class-wise training losses, we estimate the class-wise weight parameters by using a sigmoid-based function as follows:
\begin{align}
w_i = \sigma(\bar{\mathcal{L}}_{C_i};\alpha,\beta):=\frac{\beta}{1+exp(-\bar{\mathcal{L}}_{C_i}/\alpha)}-\frac{\beta}{2},
\end{align}
where $\alpha$ and $\beta$ are the parameters to determine the shape of the sigmoid-based function, which are optimized automatically.
From the derivation, we can find that $\sigma(\bar{\mathcal{L}_{C_i}};\alpha,\beta)$ ranges from $-\beta/2$ to $\beta/2$.
$\alpha$ determines the magnitude of the gradient, and $\beta$ changes the boundary scale for the class-wise training losses.

To optimize the values of $\alpha$ and $\beta$, we design a new objective function to reduce the variance of the class-wise training losses as follows:
\begin{equation}
\begin{aligned}
    &\mathcal{L}_{\sigma(\alpha, \beta)}=\eta(\mathcal{L}_{C}^1)\approx\\
    &\sqrt{\sum_{i=1}^{c}\bigg({\mathcal{L}_{C_i}^0-\sigma(\bar{\mathcal{L}}_{C_i}^1;\alpha,\beta)\cdot (\mathcal{L}_{C_i}^1 - \mathcal{L}_{C_i}^0)-\mathbb{E}[\mathcal{L}_C^1]}\bigg)^2},
\end{aligned}
\end{equation}
where $\eta(\mathcal{L}_{C}^1)$ is a function for a standard deviation of class-wise training losses after the first iteration, and $\mathcal{L}_{C_i}^0$ and $\mathcal{L}_{C_i}^1$ mean the $i$-th class-wise training losses before and after the first training iteration, respectively.
Thus, we approximate the variance of the class-wise training losses at the current iteration by using a first-order Taylor series to simplify the objective function.
We use the \textit{SLSQP} optimizer~\cite{perez2012pyopt} to find the optimal solution of $\alpha$ and $\beta$, and we freeze the optimal values during the entire training sequence for the calibration scaling.

Now we use the optimal alpha and beta to estimate the class-wise loss scaling weights and calculate the weighted sum of class-wise loss with class-wise loss scaling weights. Thus, we can define $i$-th class-wise loss scaling as follows:
\begin{equation}
\mathcal{L}_{P_i} = w_i\mathcal{L}_{C_i}.
\label{Eq:11}
\end{equation}

Based on empirical comparisons provided in Appendix~\ref{Appendix:A}, we have decided to employ a normalization method based on the normal distribution among the various normalization methods.
%In Appendix~\ref{Appendix:A}, we provide the results of a comparison with various normalization methods and explain the motivation for selecting a normal distribution.

\subsection{Calibration Scaling}

Finally, from the definition of class-wise training losses and class-wise loss scaling, we optimize the calibration function scaling by the combination of class-wise training losses and class-wise loss scaling. 
The total loss is $ \mathcal{L}_{T} = \sum_{i=1}^{c} \mathcal{L}_{C_i} + \mathcal{L}_{P_i}$, which can be simplified by Eq.~\ref{Eq:11} as follows:
\begin{equation}
    \mathcal{L}_{T} = \sum_{i=1}^{c} (1 + w_i)\mathcal{L}_{C_i}.
\end{equation}

To find optimal parameters, we utilize the original optimizer and hyperparameters of baselines.

\begin{figure*}
    \centering
    \includegraphics[width=0.95\linewidth]{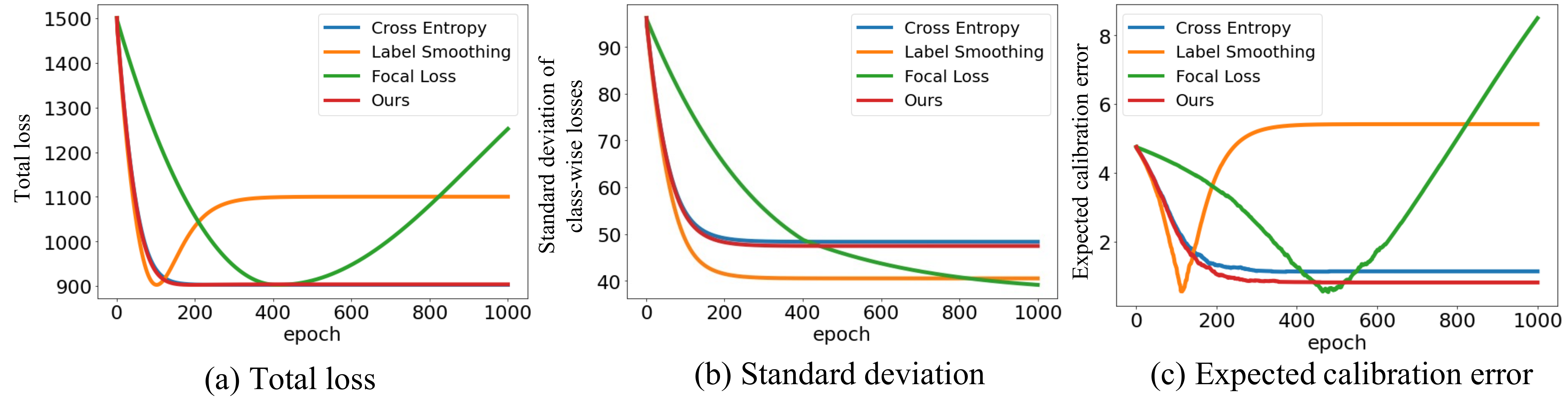}
\caption{\label{fig:generalization}\textbf{Training stability of losses.} (a), (b) and (c) show the total loss, the standard deviation of class-wise losses, and the expected calibration error according to the training epoch. While \textit{Label Smoothing} and \textit{Focal Loss} are diverging from specific training epochs, our loss is stably aligned with the conventional cross-entropy loss.}
\end{figure*}

\begin{table*}[t]
\caption{\label{tab:comp}The comparison of calibration performance on ECE (Accuracy) for various image classification datasets and models. Above one is performance and below one are gain or loss. Bold is the state-of-the-art performance for each, and blue and red colors respectively indicate improvement and degrade compared with the baseline score.}
\label{ablation}
\resizebox{\textwidth}{!}{%
\begin{tabular}{c c || c c c c | c c c c | c c }
\hline
\multicolumn{2}{c||}{\multirow{2}{*}{Method}} & \multicolumn{4}{c|}{CIFAR10} & \multicolumn{4}{c|}{CIFAR100} & \multicolumn{2}{c}{ImageNet} \\

    & & DenseNet40 & WideResNet32 & ResNet110 & ResNet110SD & DenseNet40 & WideResNet32 & ResNet110 & ResNet110SD & ResNet152 & DenseNet161  \\
\hline\hline
&Uncalibrated                          & 5.50 (92.42) &  4.51 (93.93) &  4.75 (93.56) &  4.11 (94.04) 
                            & 21.16 (70.00) & 18.78 (73.82) & 18.48 (71.48) & 15.86 (72.83) 
                            & 6.54 (76.20) &  5.72 (77.05) \\
\hline\hline
\multirow{5}{*}{\rotatebox{270}{Baseline~~}}
&TS~\cite{TS}               & 0.96 (92.42) &  0.82 (93.93) &  1.14 (93.56) &  0.54 (94.04) 
                            & 0.97 (70.00) &  1.49 (73.82) &  2.47 (71.48) &  1.27 (72.83) 
                            & 2.07 (76.20) &  1.96 (77.05) \\
&ETS~\cite{ETS}             & 0.97 (92.42) &  0.81 (93.93) &  1.14 (93.56) &  0.54 (94.04) 
                            & 0.99 (70.00) &  1.50 (73.82) &  2.50 (71.48) &  1.25 (72.83) 
                            & 2.06 (76.20) &  1.96 (77.05) \\
&PTS~\cite{PTS}             & 0.95 (92.42) &  0.78 (93.93) &  1.13 (93.56) &  0.56 (94.04) 
                            & 0.90 (70.00) &  1.47 (73.82) &  2.38 (71.48) &  1.21 (72.83) 
                            & 2.08 (76.20) &  1.94 (77.05) \\
&CTS~\cite{CTS}             & 1.01 (92.48) &  0.84 (94.21) &  1.31 (93.43) &  0.75 (94.20) 
                            & 0.99 (70.22) &  1.61 (74.03) &  2.50 (71.39) &  1.15 (73.51) 
                            & 2.38 (75.98) &  2.24 (76.78) \\
\cline{2-12}
&Average                    & 0.97 (92.44) &  0.82 (\textbf{94.00}) &  1.18 (93.53) &  0.60 (\textbf{94.08}) 
                            & 0.96 (70.06) &  1.52 (73.87) &  2.46 (71.46) &  1.22 (73.00) 
                            & 2.15 (76.15) &  2.03 (76.98) \\
                         
\hline\hline
\multirow{10}{*}{\rotatebox{270}{Label Smoothing (LS)~~}\rotatebox{270}{\multirow{4}{*}{~~\cite{muller2019does}}}}
&\multirow{2}{*}{TS+LS}     &  0.44 (92.42) &  0.68 (93.93) &  1.20 (93.56) &  1.29 (94.04) 
                            &  3.26 (70.00) &  3.15 (73.82) &  1.97 (71.48) &  2.38 (72.83) 
                            &  6.75 (76.20) &  6.55 (77.05) \\
                            
                            && ({\small{\color{blue}{-0.52}} (0.00)}) & ({\small{\color{blue}{-0.14}} (0.00)}) & ({\small{\color{red}{+0.06}} (0.00)}) & ({\small{\color{red}{+0.75}} (0.00)})
                            & ({\small{\color{red}{+2.29}} (0.00)})& ({\small{\color{red}{+1.66}} (0.00)}) & ({\small{\color{blue}{-0.50}} (0.00)}) & ({\small{\color{red}{+1.11}} (0.00)})
                            & ({\small{\color{red}{+4.68}} (0.00)}) & ({\small{\color{red}{+4.59}} (0.00)})\\

&\multirow{2}{*}{ETS+LS}    &  8.27 (92.42) &  7.35 (93.93) &  5.13 (93.56) &  8.59 (94.04) 
                            &  8.94 (70.00) &  7.10 (73.82) &  5.29 (71.48) &  9.57 (72.83) 
                            &  8.19 (76.20) &  7.84 (77.05) \\
                            
                            && ({\small{\color{red}{+7.30}} (0.00)}) & ({\small{\color{red}{+6.54}} (0.00)}) & ({\small{\color{red}{+3.99}} (0.00)}) & ({\small{\color{red}{+8.05}} (0.00)})
                            & ({\small{\color{red}{+7.95}} (0.00)})& ({\small{\color{red}{+5.60}} (0.00)}) & ({\small{\color{red}{+2.79}} (0.00)}) & ({\small{\color{red}{+8.32}} (0.00)})
                            & ({\small{\color{red}{+6.13}} (0.00)}) & ({\small{\color{red}{+5.88}} (0.00)})\\

&\multirow{2}{*}{PTS+LS}    &  8.73 (92.42) &  7.76 (93.93) &  5.42 (93.56) &  9.03 (94.04) 
                            &  9.61 (70.00) &  7.68 (73.82) &  5.86 (71.48) &  10.12 (72.83) 
                            &  8.48 (76.20) &  8.06 (77.05) \\
                            
                            && ({\small{\color{red}{+7.78}} (0.00)}) & ({\small{\color{red}{+6.98}} (0.00)}) & ({\small{\color{red}{+4.29}} (0.00)}) & ({\small{\color{red}{+8.47}} (0.00)})
                            & ({\small{\color{red}{+8.71}} (0.00)})& ({\small{\color{red}{+6.21}} (0.00)}) & ({\small{\color{red}{+3.48}} (0.00)}) & ({\small{\color{red}{+8.91}} (0.00)})
                            & ({\small{\color{red}{+6.40}} (0.00)}) & ({\small{\color{red}{+6.12}} (0.00)})\\
                            
&\multirow{2}{*}{CTS+LS}    &  1.21 (92.38) &  1.13 (94.09) &  0.63 (93.49) &  0.52 (94.08) 
                            &  3.29 (70.27) &  3.20 (74.04) &  2.17 (71.55) &  2.13 (73.52) 
                            &  6.79 (76.15) &  5.84 (77.01) \\

                            && ({\small{\color{red}{+0.20}} ({\color{red}-0.09})}) & ({\small{\color{red}{+0.29}} ({\color{red}-0.12})}) & ({\small{\color{blue}{-0.68}} ({\color{blue}-0.06})}) & ({\small{\color{blue}{-0.23}} ({\color{red}-0.12})})
                            & ({\small{\color{blue}{-2.30}} ({\color{blue}+0.05})})& ({\small{\color{red}{+1.59}} ({\color{blue}+0.01})}) & ({\small{\color{blue}{-0.33}} ({\color{blue}+0.16})}) & ({\small{\color{red}{+0.98}} ({\color{blue}+0.01})})
                            & ({\small{\color{red}{+4.41}} ({\color{blue}+0.17})}) & ({\small{\color{red}{+3.60}} ({\color{blue}+0.23})})\\

\cline{2-12}
&\multirow{2}{*}{Average}   & 4.66 (92.41) &  4.23 (93.97) &  3.10 (93.54) &  4.86 (94.05) 
                            & 6.28 (70.07) &  5.28 (\textbf{73.88}) &  3.82 (\textbf{71.50}) &  6.05 (73.00) 
                            & 7.55 (\textbf{76.19}) &  7.07 (77.04) \\
                            
                            && ({\small{\color{red}{+3.69}} ({\color{red}-0.03})}) & ({\small{\color{red}{+3.41}} ({\color{red}-0.03})}) & ({\small{\color{red}{+1.92}} ({\color{blue}+0.01})}) & ({\small{\color{red}{+4.24}} ({\color{red}-0.03})})
                            & ({\small{\color{red}{+5.32}} ({\color{blue}+0.01})})& ({\small{\color{red}{+3.76}} ({\color{blue}+0.01})}) & ({\small{\color{red}{+1.36}} ({\color{blue}+0.04})}) & ({\small{\color{red}{+4.83}} ({0.00})})
                            & ({\small{\color{red}{+5.40}} ({\color{blue}+0.04})}) & ({\small{\color{red}{+5.04}} ({\color{blue}+0.06})})\\
                            
\hline\hline
\multirow{10}{*}{\rotatebox{270}{~~~~Focal Loss (FL)}\rotatebox{270}{\multirow{4}{*}{\cite{mukhoti2020calibrating}~~~}}}
&\multirow{2}{*}{TS+FL}     &  1.75 (92.42) &  1.46 (93.93) &  0.58 (93.56) &  0.84 (94.04) 
                            &  3.45 (70.00) &  3.20 (73.82) &  1.67 (71.48) &  2.30 (72.83) 
                            &  4.70 (76.20) &  4.73 (77.05) \\
                            
                            && ({\small{\color{red}{+0.79}} (0.00)}) & ({\small{\color{red}{+0.64}} (0.00)}) & ({\small{\color{blue}{-0.56}} (0.00)}) & ({\small{\color{red}{+0.30}} (0.00)})
                            & ({\small{\color{red}{+2.48}} (0.00)})& ({\small{\color{red}{+1.71}} (0.00)}) & ({\small{\color{blue}{-0.80}} (0.00)}) & ({\small{\color{red}{+1.03}} (0.00)})
                            & ({\small{\color{red}{+2.63}} (0.00)}) & ({\small{\color{red}{+2.77}} (0.00)})\\

&\multirow{2}{*}{ETS+FL}    &  8.94 (92.42) &  11.27 (93.93) &  14.72 (93.56) &  8.33 (94.04) 
                            &  11.07 (70.00) &  9.36 (73.82) &  10.46 (71.48) &  10.52 (72.83) 
                            &  7.59 (76.20) &  7.78 (77.05) \\
                            
                            && ({\small{\color{red}{+7.97}} (0.00)}) & ({\small{\color{red}{+10.46}} (0.00)}) & ({\small{\color{red}{+13.58}} (0.00)}) & ({\small{\color{red}{+7.79}} (0.00)})
                            & ({\small{\color{red}{+10.08}} (0.00)})& ({\small{\color{red}{+7.86}} (0.00)}) & ({\small{\color{red}{+7.96}} (0.00)}) & ({\small{\color{red}{+9.27}} (0.00)})
                            & ({\small{\color{red}{+5.53}} (0.00)}) & ({\small{\color{red}{+5.82}} (0.00)})\\

&\multirow{2}{*}{PTS+FL}    &  9.62 (92.42) &  11.88 (93.93) &  15.68 (93.56) &  8.99 (94.04) 
                            &  11.91 (70.00) &  10.12 (73.82) &  11.10 (71.48) &  11.07 (72.83)
                            &  7.78 (76.20) &  8.05 (77.05) \\
                            
                            && ({\small{\color{red}{8.67}} (0.00)}) & ({\small{\color{red}{11.1}} (0.00)}) & ({\small{\color{red}{+14.55}} (0.00)}) & ({\small{\color{red}{+8.43}} (0.00)})
                            & ({\small{\color{red}{+11.01}} (0.00)})& ({\small{\color{red}{+8.65}} (0.00)}) & ({\small{\color{red}{+8.72}} (0.00)}) & ({\small{\color{red}{+9.86}} (0.00)})
                            & ({\small{\color{red}{+5.70}} (0.00)}) & ({\small{\color{red}{+6.11}} (0.00)})\\
                            
&\multirow{2}{*}{CTS+FL}    &  1.76 (92.37) &  1.47 (93.95) &  0.81 (93.58) &  0.83 (94.07) 
                            &  3.87 (70.32) &  3.94 (74.00) &  1.75 (71.56) &  1.15 (73.53) 
                            &  1.83 (76.12) &  1.94 (77.08) \\
                            
                            && ({\small{\color{red}{+0.75}} (\color{red}-0.11)}) & ({\small{\color{red}{+0.63}} (\color{red}-0.26)}) & ({\small{\color{blue}{-0.05}} (\color{blue}+0.15)}) & ({\small{\color{red}{+0.08}} (\color{red}-0.13)})
                            & ({\small{\color{red}{+2.88}} (\color{blue}+0.10)})& ({\small{\color{red}{+2.33}} (\color{red}-0.03)}) & ({\small{\color{blue}{-0.75}} (\color{blue}+0.17)}) & ({\small{{0.00}} (\color{blue}+0.02)})
                            & ({\small{\color{blue}{-0.55}} (\color{blue}+0.14)}) & ({\small{\color{blue}{-0.30}} (\color{blue}+0.30)})\\

\cline{2-12}
\multirow{10}{*}{\rotatebox{270}{~~~~~~~~~~~~~~~~~~~~Ours}}
&\multirow{2}{*}{Average}   & 5.52 (92.41) &  6.52 (93.94) &  7.95 (\textbf{93.57}) &  4.75 (94.05) 
                            & 7.58 (\textbf{70.08}) &  6.66 (73.87) &  6.25 (\textbf{71.50}) &  6.26 (\textbf{73.01}) 
                            & 5.48 (76.18) &  5.63 (\textbf{77.06}) \\
                            
                            && ({\small{\color{red}{+4.55}} ({\color{red}-0.03})}) & ({\small{\color{red}{+5.70}} ({\color{red}-0.06})}) & ({\small{\color{red}{+6.77}} ({\color{blue}+0.04})}) & ({\small{\color{red}{+4.15}} ({\color{red}-0.03})})
                            & ({\small{\color{red}{+6.62}} ({\color{blue}+0.02})})& ({\small{\color{red}{+5.14}} ({0.00})}) & ({\small{\color{red}{+3.79}} ({\color{blue}+0.04})}) & ({\small{\color{red}{+5.04}} ({0.00})})
                            & ({\small{\color{red}{+3.33}} ({\color{blue}+0.03})}) & ({\small{\color{red}{+3.60}} ({\color{blue}+0.08})})\\
                                                        
\hline\hline
&\multirow{2}{*}{TS+Ours}   &  0.71 (92.42) &  0.78 (93.93) &  0.85 (93.56) &  0.49 (94.04) 
                            &  0.81 (70.00) &  1.23 (73.82) &  1.84 (71.48) &  1.18 (72.83) 
                            &  2.01 (76.20) &  1.93 (77.05) \\
                            
                            && ({\small{\color{blue}{-0.25}} (0.00)}) & ({\small{\color{blue}{-0.04}} (0.00)}) & ({\small{\color{blue}{-0.29}} (0.00)}) & ({\small{\color{blue}{-0.05}} (0.00)})
                            & ({\small{\color{blue}{-0.16}} (0.00)}) & ({\small{\color{blue}{-0.26}} (0.00)}) & ({\small{\color{blue}{-0.63}} (0.00)}) & ({\small{\color{blue}{-0.09}} (0.00)})
                            & ({\small{\color{blue}{-0.06}} (0.00)}) & ({\small{\color{blue}{-0.03}} (0.00)})\\

&\multirow{2}{*}{ETS+Ours}  &  0.71 (92.42) &  0.76 (93.93) &  0.86 (93.56) &  0.47 (94.04) 
                            &  0.85 (70.00) &  1.15 (73.82) &  1.85 (71.48) &  1.02 (72.83) 
                            &  2.02 (76.20) &  1.96 (77.05) \\
                            
                            && ({\small{\color{blue}{-0.26}} (0.00)}) & ({\small{\color{blue}{-0.05}} (0.00)}) & ({\small{\color{blue}{-0.28}} (0.00)}) & ({\small{\color{blue}{-0.07}} (0.00)})
                            & ({\small{\color{blue}{-0.14}} (0.00)}) & ({\small{\color{blue}{-0.35}} (0.00)}) & ({\small{\color{blue}{-0.65}} (0.00)}) & ({\small{\color{blue}{-0.23}} (0.00)})
                            & ({\small{\color{blue}{-0.03}} (0.00)}) & ({\small{{0.00}} (0.00)})\\

&\multirow{2}{*}{PTS+Ours}  &  0.71 (92.42) &  0.77 (93.93) &  0.82 (93.56) &  0.47 (94.04) 
                            &  0.80 (70.00) &  1.18 (73.82) &  1.88 (71.48) &  1.05 (72.83) 
                            &  2.03 (76.20) &  1.92 (77.05) \\
                            
                            && ({\small{\color{blue}{-0.24}} (0.00)}) & ({\small{\color{blue}{-0.01}} (0.00)}) & ({\small{\color{blue}{-0.31}} (0.00)}) & ({\small{\color{blue}{-0.09}} (0.00)})
                            & ({\small{\color{blue}{-0.10}} (0.00)}) & ({\small{\color{blue}{-0.29}} (0.00)}) & ({\small{\color{blue}{-0.50}} (0.00)}) & ({\small{\color{blue}{-0.16}} (0.00)})
                            & ({\small{\color{blue}{-0.05}} (0.00)}) & ({\small{\color{blue}{-0.02}} (0.00)})\\
                            
&\multirow{2}{*}{CTS+Ours}  &  0.74 (92.60) &  0.44 (94.09) &  1.04 (93.45) &  0.61 (94.14) 
                            &  0.74 (70.23) &  1.42 (73.86) &  1.99 (71.41) &  0.97 (73.42) 
                            &  1.68 (75.91) &  1.60 (76.67) \\
                            
                            && ({\small{\color{blue}{-0.27}} (\color{blue}+0.12)}) & ({\small{\color{blue}{-0.40}} (\color{red}-0.12)}) & ({\small{\color{blue}{-0.27}} (\color{blue}+0.02)}) & ({\small{\color{blue}{-0.14}} (\color{red}-0.06)})
                            & ({\small{\color{blue}{-0.25}} (\color{blue}+0.01)}) & ({\small{\color{blue}{-0.19}} (\color{red}-0.17)}) & ({\small{\color{blue}{-0.51}} (\color{blue}+0.02)}) & ({\small{\color{blue}{-0.18}} (\color{red}-0.09)})
                            & ({\small{\color{blue}{-0.70}} (\color{red}-0.07)}) & ({\small{\color{blue}{-0.64}} (\color{red}-0.11)})\\

\cline{2-12}
&\multirow{2}{*}{Average}    & \textbf{0.72} (\textbf{92.47}) &  \textbf{0.69} (93.97) &  \textbf{0.89} (93.53) &  \textbf{0.51} (94.07)
                            & \textbf{0.80} (70.06) &  \textbf{1.25} (73.83) &  \textbf{1.89} (71.46) & \textbf{1.06} (72.98) 
                            & \textbf{1.94} (76.13) &  \textbf{1.85} (76.96) \\
                            
                            && ({\small{\color{blue}{-0.25}} (\color{blue}+0.03)}) & ({\small{\color{blue}{-0.12}} (\color{red}-0.03)}) & ({\small{\color{blue}{-0.29}} (0.00)}) & ({\small{\color{blue}{-0.09}} (\color{red}-0.01)})
                            & ({\small{\color{blue}{-0.16}} (0.00)}) & ({\small{\color{blue}{-0.27}} (\color{red}-0.04)}) & ({\small{\color{blue}{-0.57}} (0.00)}) & ({\small{\color{blue}{-0.16}} (\color{red}-0.02)})
                            & ({\small{\color{blue}{-0.21}} (\color{red}-0.02)}) & ({\small{\color{blue}{-0.18}} (\color{red}-0.02)})\\\hline
\end{tabular}}
\end{table*}

\section{Experiments}
In this section, we present the empirical results on diverse classifier sets.
Then, we describe the results of our extensive analysis and validate our approach compared to other methods.

\subsection{Dataset and Implementation Details}
We use three different datasets and six different pre-trained models to train and evaluate the calibration methods. We separate the datasets into validation datasets and test datasets with the sizes of 25000/10000 for CIFAR10 and CIFAR100 datasets~\cite{Krizhevsky_2009_17719} and 25000/25000 for ImageNet dataset~\cite{deng2009imagenet}. 
For the imbalanced dataset experiment, we separate the datasets into train, validation, and test datasets with the sizes of 10215/10216/10000 for CIFAR10-LT and 9786/9787/10000 for CIFAR100-LT~\cite{tang2020longtailed}.
In these datasets, the number of classes varies from 10 to 1000. Each post-hoc calibration method was trained with validation datasets and evaluated by test datasets. We use pre-trained but uncalibrated models, which are DenseNet40,DenseNet161~\cite{huang2017densely}, WideResNet32~\cite{BMVC2016_87}, ResNet110, ResNet152~\cite{7780459}, and ResNet110SD~\cite{10.1007/978-3-319-46493-0_39}. We follow the datasets and the pre-trained model setting referred to in \cite{MetaCal} and use the logits, which are generated by combinations of datasets and models, as their original code\footnote{\tiny{\url{https://github.com/markus93/NN\_calibration}}}.
We initialize $\alpha$ and $\beta$ by 1.0 and 1.5, respectively, before their optimization.
We conduct our experiments upon one RTX3090 environment. Our code is available online\footnote{\tiny\url{https://github.com/SeungjinJung/SCTL}}.
% Our code is available for access at the URL
\subsection{Evaluation Measure}
We utilize two evaluation measures to compare the proposed algorithm with the previous methods.
First, the accuracy measure is used to compare the model classification performance.
The performance gap between the baseline and the calibrated model should be small for the successful calibration.
Second, we use the Expected Calibration Error (ECE)~\cite{ECE} to measure the calibration error as an indicator of the model's confidence.
The ECE measures the discrepancy between the accuracy and the predicted probability within a given probability interval.
Therefore, the ECE value goes small for the well-calibrated model.

\begin{table*}[t]
\caption{\label{tab:imcomp}The comparison of calibration results on ECE (Accuracy) for several imbalanced datasets and models for image classification. Performance is written at an upper line, while gain or loss is at a lower line. Each dataset's and model's best performance is displayed in bold, while blue and red digits denote improvement and degradation compared to the baseline score, respectively.}
\label{ablation}
\resizebox{\textwidth}{!}{%
\begin{tabular}{c c || c c c c | c c c c }
\hline
\multicolumn{2}{c||}{\multirow{2}{*}{Method}} & \multicolumn{4}{c|}{CIFAR10-LT} & \multicolumn{4}{c}{CIFAR100-LT} \\
    % \cline{2-11}
    && DenseNet40 & WideResNet28 & ResNet110 & ResNet110SD & DenseNet40 & WideResNet28 & ResNet110 & ResNet110SD\\
\hline\hline
&Uncalibrated               & 21.08 (72.11) & 9.11 (85.38) & 19.14 (75.97) & 14.43 (64.90) 
                            & 38.40 (37.97) & 9.11 (53.57) & 41.98 (38.57) & 30.32 (34.87) \\
\hline\hline
\multirow{5}{*}{\rotatebox{270}{Baseline~~}}
&TS~\cite{TS}                & 6.16 (72.11) & 4.47 (85.38) & 5.94 (75.97) & 6.84 (64.90) 
                            & 8.89 (37.97) & 9.62 (53.57) & 8.46 (38.57) & 4.81 (34.87) \\
&ETS~\cite{ETS}              & 6.15 (72.11) & 4.46 (85.38) & 5.93 (75.97) & 6.84 (64.90) 
                            & 8.85 (37.97) & 9.60 (53.57) & 8.38 (38.57) & 4.76 (34.87) \\
&PTS~\cite{PTS}              & 6.10 (72.11) & 4.43 (85.38) & 5.90 (75.97) & 6.76 (64.90) 
                            & 8.77 (37.97) & 9.60 (53.57) & 8.29 (38.57) & 4.69 (34.87) \\
&CTS~\cite{CTS}              & 6.02 (72.37) & 4.16 (86.42) & 5.74 (76.17) & 7.36 (64.10) 
                            & 11.54 (36.72) & 8.06 (56.03) & 11.43 (36.55) & 7.43 (33.27)  \\
\cline{2-10}
&Average                    & 6.11 (72.18) & 4.38 (85.64) & 5.88 (76.02) & 6.95 (64.70)
                            & 9.51 (37.66) & 9.22 (54.19) & 9.14 (38.07) & 5.42 (34.47) \\
\hline\hline
\multirow{10}{*}{\rotatebox{270}{Label Smoothing (LS)~~}\rotatebox{270}{\multirow{4}{*}{~\cite{muller2019does}}}}
&\multirow{2}{*}{TS+LS}     & 8.16 (72.11) & 2.24 (85.38) & 8.99 (75.97) & 2.90 (64.90) 
                            & 12.37 (37.97) & 7.43(53.57) & 15.02 (38.57) & 5.85 (34.87) \\
                            && ({\small{\color{red}{+2.00}} (0.00)}) & ({\small{\color{blue}{-2.23}} (0.00)}) & ({\small{\color{red}{+3.05}} (0.00)}) & ({\small{\color{blue}{-3.94}} (0.00)})
                            & ({\small{\color{red}{+3.48}} (0.00)})& ({\small{\color{blue}{-2.19}} (0.00)}) & ({\small{\color{red}{+6.56}} (0.00)}) & ({\small{\color{red}{+1.04}} (0.00)})\\
&\multirow{2}{*}{ETS+LS}    & 0.74 (72.11) & 1.54 (85.38) & 1.49 (75.97) & 1.63 (64.90) 
                            & 4.40 (37.97) & 6.89 (53.57) & 4.64 (38.57) & 1.74 (34.87) \\
                            && ({\small{\color{blue}{-5.41}} (0.00)}) & ({\small{\color{blue}{-2.92}} (0.00)}) & ({\small{\color{blue}{-4.44}} (0.00)}) & ({\small{\color{blue}{-5.21}} (0.00)})
                            & ({\small{\color{blue}{-4.45}} (0.00)}) & ({\small{\color{blue}{-2.71}} (0.00)}) & ({\small{\color{blue}{-3.74}} (0.00)}) & ({\small{\color{blue}{-3.02}} (0.00)})\\
&\multirow{2}{*}{PTS+LS}    & 1.02 (72.11) & 1.67 (85.38) & 1.12 (75.97) & 1.50 (64.90) 
                            & 3.62 (37.97) & 6.72 (53.57) & 3.85 (38.57) & 1.54 (34.87) \\
                            && ({\small{\color{blue}{-5.08}} (0.00)}) & ({\small{\color{blue}{-2.76}} (0.00)}) & ({\small{\color{blue}{-4.78}} (0.00)}) & ({\small{\color{blue}{-5.26}} (0.00)})
                            & ({\small{\color{blue}{-5.15}} (0.00)}) & ({\small{\color{blue}{-2.88}} (0.00)}) & ({\small{\color{blue}{-4.44}} (0.00)}) & ({\small{\color{blue}{-3.15}} (0.00)})\\
&\multirow{2}{*}{CTS+LS}    & 8.13 (72.00) & 2.59 (86.03) & 8.94 (76.08) & 4.59 (63.76)
                            & 14.35 (37.27) & 6.76 (53.75) & 16.94 (37.77) & 8.47 (33.64) \\
                            && ({\small{\color{red}{+2.11}} (\color{red}-0.37)}) & ({\small{\color{blue}{-1.57}} (\color{red}-0.39)}) & ({\small{\color{red}{3.20}} (\color{red}-0.09)}) & ({\small{\color{blue}{-2.77}} (\color{red}-0.34)})
                            & ({\small{\color{red}{+2.81}} (\color{blue}+0.55)}) & ({\small{\color{blue}{-1.30}} (\color{red}-0.28)}) & ({\small{\color{red}{+5.51}} (\color{blue}+1.22)}) & ({\small{\color{red}{+1.04}} (\color{blue}+0.37)})\\
\cline{2-10}
&\multirow{2}{*}{Average}   & 4.51 (72.08) & \textbf{2.01} (85.54) & 5.14 (76.00) & 2.66 (64.62)
                            & 8.69 (37.80) & 6.95 (53.62) & 10.11 (38.37) & 4.40 (34.56)\\
                            && ({\small{\color{blue}{-1.60}} (\color{red}-0.10)}) & ({\small{\color{blue}{-2.37}} (\color{red}-0.10)}) & ({\small{\color{blue}{-0.74}} (\color{red}-0.02)}) & ({\small{\color{blue}{-4.29}} (\color{red}-0.08)})
                            & ({\small{\color{blue}{-0.82}} (\color{blue}+0.14)}) & ({\small{\color{blue}{-2.27}} (\color{red}-0.57)}) & ({\small{\color{red}{+0.97}} (\color{blue}+0.30)}) & ({\small{\color{blue}{-1.02}} (\color{blue}+0.09)})\\
\hline\hline
\multirow{10}{*}{\rotatebox{270}{~~~~~Focal Loss (FL)}\rotatebox{270}{\multirow{4}{*}{\cite{mukhoti2020calibrating}~~~}}}
&\multirow{2}{*}{TS+FL}     & 7.24 (72.11) & 1.68 (85.38) & 8.45 (75.97) & 4.33 (64.90)
                            & 10.01 (37.97) & 6.04 (53.57) & 12.91 (38.57) & 2.31 (34.87) \\
                            && ({\small{\color{red}{+1.08}} (0.00)}) & ({\small{\color{blue}{-2.79}} (0.00)}) & ({\small{\color{red}{+2.51}} (0.00)}) & ({\small{\color{blue}{-2.51}} (0.00)})
                            & ({\small{\color{red}{+1.12}} (0.00)})& ({\small{\color{blue}{-3.58}} (0.00)}) & ({\small{\color{red}{+4.45}} (0.00)}) & ({\small{\color{blue}{-2.50}} (0.00)})\\
&\multirow{2}{*}{ETS+FL}    & 9.57 (72.11) & 11.69 (85.38) & 9.51 (75.97) & 9.85 (64.90)
                            & 0.96 (37.97) & 6.15 (53.57) & 1.24 (38.57) & 3.56 (34.87) \\
                            && ({\small{\color{red}{+3.42}} (0.00)}) & ({\small{\color{red}{+7.23}} (0.00)}) & ({\small{\color{red}{+3.58}} (0.00)}) & ({\small{\color{red}{+3.01}} (0.00)})
                            & ({\small{\color{blue}{-7.89}} (0.00)})& ({\small{\color{blue}{-3.45}} (0.00)}) & ({\small{\color{blue}{-7.14}} (0.00)}) & ({\small{\color{blue}{-1.20}} (0.00)})\\
&\multirow{2}{*}{PTS+FL}    & 10.52 (72.11) & 12.42 (85.38) & 10.20 (75.97) & 10.24 (64.90)
                            & 1.25 (37.97) & 6.17 (53.57) & 1.76 (38.57) & 4.53 (34.87) \\
                            && ({\small{\color{red}{+4.42}} (0.00)}) & ({\small{\color{red}{+7.99}} (0.00)}) & ({\small{\color{red}{+4.30}} (0.00)}) & ({\small{\color{red}{+3.48}} (0.00)})
                            & ({\small{\color{blue}{-7.52}} (0.00)})& ({\small{\color{blue}{-3.43}} (0.00)}) & ({\small{\color{blue}{-6.53}} (0.00)}) & ({\small{\color{blue}{-0.16}} (0.00)})\\
&\multirow{2}{*}{CTS+FL}    &  9.82 (72.12) &  1.35 (86.01) &  10.58 (76.16) & 2.60 (65.26) 
                            &  11.26 (37.78) &  5.40 (55.86) &  14.48 (38.10) &  4.52 (34.06) \\
                            && ({\small{\color{red}{+3.80}} ({\color{red}-0.25})}) & ({\small{\color{blue}{-2.81}} ({\color{red}-0.41})}) & ({\small{\color{red}{+4.84}} ({\color{red}-0.01})}) & ({\small{\color{blue}{-4.76}} ({\color{blue}-1.16})})
                            & ({\small{\color{blue}{-0.28}} ({\color{blue}+1.06})})& ({\small{\color{blue}{-2.66}} ({\color{red}-0.17})}) & ({\small{\color{red}{+3.05}} ({\color{blue}+1.55})}) & ({\small{\color{blue}{-2.91}} ({\color{blue}+0.79})})\\
\cline{2-10}
&\multirow{2}{*}{Average}   &  9.29 (72.11) &  6.79 (85.54) &  9.69 (76.02) & 6.76 (64.99) 
                            &  5.87 (37.92) &  \textbf{5.94} (54.14) &  7.60 (38.45) & 3.73 (34.67) \\
                            && ({\small{\color{red}{+3.18}} ({\color{red}-0.07})}) & ({\small{\color{red}{+2.41}} ({\color{red}-0.10})}) & ({\small{\color{red}{+3.81}} (0.00)}) & ({\small{\color{blue}{-0.19}} ({\color{blue}+0.29})})
                            & ({\small{\color{blue}{-3.64}} ({\color{blue}+0.27})})& ({\small{\color{blue}{-3.28}} ({\color{red}-0.05})}) & ({\small{\color{blue}{-1.54}} ({\color{blue}+0.38})}) & ({\small{\color{blue}{-1.69}} ({\color{blue}+0.20})})\\
\hline\hline
\multirow{10}{*}{\rotatebox{270}{Ours~~~}}
&\multirow{2}{*}{TS+Ours}   & 3.22 (72.11) & 3.19 (85.38) & 3.53 (75.97) & 2.53 (64.90)
                            & 4.44 (37.97) & 6.99 (53.57) & 4.50 (38.57) & 1.48 (34.87) \\
                            && ({\small{\color{blue}{-2.94}} (0.00)}) & ({\small{\color{blue}{-1.28}} (0.00)}) & ({\small{\color{blue}{-2.41}} (0.00)}) & ({\small{\color{blue}{-4.31}} (0.00)})
                            & ({\small{\color{blue}{-4.45}} (0.00)})& ({\small{\color{blue}{-2.63}} (0.00)}) & ({\small{\color{blue}{-3.96}} (0.00)}) & ({\small{\color{blue}{-3.33}} (0.00)})\\
&\multirow{2}{*}{ETS+Ours}  & 3.08 (72.11) & 3.17 (85.38) & 3.40 (75.97) & 2.39 (64.90)
                            & 4.18 (37.97) & 6.92 (53.57) & 4.23 (38.57) & 1.56 (34.87) \\
                            && ({\small{\color{blue}{-3.07}} (0.00)}) & ({\small{\color{blue}{-1.29}} (0.00)}) & ({\small{\color{blue}{-2.53}} (0.00)}) & ({\small{\color{blue}{-4.45}} (0.00)})
                            & ({\small{\color{blue}{-4.67}} (0.00)})& ({\small{\color{blue}{-2.68}} (0.00)}) & ({\small{\color{blue}{-4.15}} (0.00)}) & ({\small{\color{blue}{-3.20}} (0.00)})\\
&\multirow{2}{*}{PTS+Ours}  & 3.13 (72.11) & 3.16 (85.38) & 3.47 (75.97) & 2.40 (64.90)
                            & 4.29 (37.97) & 6.90 (53.57) & 4.33 (38.57) & 1.45 (34.87) \\
                            && ({\small{\color{blue}{-2.97}} (0.00)}) & ({\small{\color{blue}{-1.27}} (0.00)}) & ({\small{\color{blue}{-2.43}} (0.00)}) & ({\small{\color{blue}{-4.36}} (0.00)})
                            & ({\small{\color{blue}{-4.48}} (0.00)}) & ({\small{\color{blue}{-2.70}} (0.00)}) & ({\small{\color{blue}{-3.96}} (0.00)}) & ({\small{\color{blue}{-3.24}} (0.00)})\\
&\multirow{2}{*}{CTS+Ours}  & 2.79 (74.99) & 1.99 (88.00) & 2.96 (78.55) & 2.19 (67.43) 
                            & 3.44 (40.13) & 5.56 (57.62) & 3.85 (39.79) & 1.50 (36.32) \\
                            && ({\small{\color{blue}{-3.23}} ({\color{blue}+2.62})}) & ({\small{\color{blue}{-2.17}} ({\color{blue}+1.58})}) & ({\small{\color{blue}{-2.78}} ({\color{blue}+2.38})}) & ({\small{\color{blue}{-5.17}} ({\color{blue}+3.33})})
                            & ({\small{\color{blue}{-8.10}} ({\color{blue}+3.41})}) & ({\small{\color{blue}{-2.50}} ({\color{blue}+1.59})}) & ({\small{\color{blue}{-7.58}} ({\color{blue}+3.24})}) & ({\small{\color{blue}{-5.93}} ({\color{blue}+3.05})})\\
\cline{2-10}
&\multirow{2}{*}{Average}   &  \textbf{3.06} (\textbf{72.83}) & 2.88 (\textbf{86.04}) & \textbf{3.34} (\textbf{76.62}) & \textbf{2.38} (\textbf{65.53})
                            &  \textbf{4.09} (\textbf{38.51}) & 6.59 (\textbf{54.58}) & \textbf{4.23} (\textbf{38.88}) & \textbf{1.50} (\textbf{35.23}) \\
                            && ({\small{\color{blue}{-3.05}} ({\color{blue}+0.65})}) & ({\small{\color{blue}{-1.50}} ({\color{blue}+0.40})}) & ({\small{\color{blue}{-2.40}} ({\color{blue}+0.60})}) & ({\small{\color{blue}{-4.57}} ({\color{blue}+0.83})})
                            & ({\small{\color{blue}{-5.42}} ({\color{blue}+0.85})}) & ({\small{\color{blue}{-2.63}} ({\color{blue}+0.32})}) & ({\small{\color{blue}{-4.91}} ({\color{blue}+0.81})}) & ({\small{\color{blue}{-3.92}} ({\color{blue}+0.76})})\\
\hline
\end{tabular}}
\end{table*}

\subsection{Comparison Result}
The detailed implementation information of our comparison models is as follows: Temperature Scaling (TS)~\cite{TS}, Ensemble Temperature Scaling (ETS)~\cite{ETS}, Class-based Temperature Scaling (CTS)~\cite{CTS}, and Parameterized Temperature Scaling (PTS)~\cite{PTS}.
The detailed implementation for the comparison methods is explained in Appendix~\ref{Appendix:B}.
We train baseline methods using a learning rate of 0.02, 1000 epochs, and a cross-entropy loss.
TS, ETS, and CTS use LBFGS optimizer~\cite{battiti1990bfgs} without a weight decay, and PTS utilizes Adam optimizer~\cite{kingma2014adam} with 0.002 weight decay.

We also compare proposed approaches with Focal Loss (FL)~\cite{mukhoti2020calibrating} and Label Smoothing (LS)~\cite{muller2019does} due to their similarity where the weights of the CE losses are scaled by multipliers.
FL uses $\gamma=3$, LS uses $\alpha=0.05$, and both use a learning rate schedule of {0.005, 0.003, 0.001} for the first 200, next 400, and last 400 epochs on the baseline method.

\subsubsection{Advantage of Balanced Class-wise Losses}
We conducted additional experiments to compare and analyze the correlation between calibration error and the balanced class-wise losses, which was found in Section 3.1.
The additional experiments were performed using the CIFAR10 dataset and ResNet110 model, and the results are presented in Fig.~\ref{fig:generalization}.
In order to find out the relationship between the class-wise losses and ECE, we use the same method to estimate the class-wise losses for all the compared algorithms according to Eq.~\ref{eq:7}.
Then, the total sum and standard deviation were calculated by using the estimated class-wise loss, as shown in Fig.~\ref{fig:generalization}-(a) and (b), respectively.
Finally, the expected calibration error is plotted in Fig.~\ref{fig:generalization}-(c).

When we compare the total loss and the expected calibration error, we can see that the trends of the total loss and the expectation calibration error are similar to each other.
Label Smoothing (LS) and Focal Loss (FL) can confirm that total loss and expected calibration error change rapidly at similar epochs, and Cross-Entropy (CE) and ours verify that the total loss and expected calibration error converge at similar epochs.
Through the results, we can see that a well-calibrated prediction can be obtained only when the overall class-wise loss is reduced.
The next thing we should pay attention to is the similar epoch of the trends.
The total loss and expected calibration error have similar trajectories but different optimization points.
CE and our method converge in the total loss, but the calibration performance increases until the standard deviation of the class-wise loss converges.
In the case of LS and FL, even if the total loss increases slightly, the calibration performance increases when the class-wise loss is balanced.
From the results, we can confirm that the balance of class-wise loss affects the calibration performance.

As shown in Fig.~\ref{fig:generalization}-(a), the FL and LS methods show that the total loss rather increases as learning continues, which is caused by overfitting on validation.
Since almost no change can be found in imbalanced samples, easy samples continue to learn less, while hard samples continue to learn more.
Because of this problem, FL and LS must use an appropriate learning schedule technique to learn a well-calibrated model, but it is almost impossible to optimize a learning schedule to generalize in all situations.
On the contrary, our method increases the relatively under-fitted loss while decreasing the over-fitted loss to balance the losses.
As a result, the training with our method becomes stable because overfitting and underfitting are compensated to each other during the model training.

\subsubsection{Quantitative Results}
We perform experiments with our method combined with prior calibration methods on various benchmark datasets.
In prior studies, FL and LS are applied only to the TS method, but we employ these approaches to other recent calibration methods such as ETS, PTS, and CTS in our experiments.
In addition, we also test our method with the imbalance datasets to verify its real-world applicability.

\textbf{Experiments on benchmark datasets.}\\
We perform the experiments using benchmark datasets first.
As shown in Table.~\ref{tab:comp}, the proposed algorithm shows the state-of-the-art performance in terms of calibration error.
Even though our method sometimes showed worse accuracy than the prior studies, the gap between the best accuracy and our performance is very small, compared to the large gap in the calibration error.
Interestingly, while the previous studies such as LS and FL show the increment of calibration error after their employment, our method shows a relatively stable enhancement of performance.
We successfully acquire the state-of-the-art calibration error even with the large-scale dataset of ImageNet where the previous algorithms suffer from the largely increased calibration error.
Furthermore, from the results using various model architectures, we can confirm that the proposed algorithm can cover the various types of architectures.
The interesting point is that we do not tune the optimizer and the hyperparameters across the various types of datasets and network architectures, which validates the impressive robustness of our method.

\textbf{Experiments on imbalanced datasets.} \\
Because our method considers the relation between class-wise losses, we also try to verify that our method performs well in class-imbalanced situations.
We make imbalanced datasets by employing the long-tailed dataset design protocol of \cite{tang2020longtailed}, which is referred to as CIFAR10-LT and CIFAR100-LT. 
The detailed information about the imbalanced datasets is explained in Appendix~\ref{Appendix:C}.
%~\footnote{\tiny\url{https://github.com/KaihuaTang/Long-Tailed-Recognition.pytorch}}

In Table.~\ref{tab:imcomp}, we show the experimental results with the same experimental setting of benchmark datasets.
Interestingly, in the imbalanced dataset, our method shows state-of-the-art performance in terms of both the calibration error and the accuracy.
Compared to the original benchmark datasets, the overall performance of baselines degrades with the enlarged calibration errors.
While the degraded performance and the large calibration error could not be much enhanced by LS and FL, the proposed algorithm can dramatically improve both the performance and calibration of the pre-trained model.
Even though FL shows an impressive enhancement in CIFAR100-LT, the lack of its generality can be found in its results of CIFAR10-LT.
However, our method shows the generality across the various types of imbalanced datasets and model architectures.
From these experiments, we can confirm that the proposed algorithm has real-world applicability for imbalanced domain environments.

\section{Conclusion}
The varying amounts of intra-class and inter-class appearance variation frequently lead the class-wise training losses to diverge, and we discover that this is what causes the uncalibrated prediction of deep neural network.
To solve the issue, we provided the calibration method to synchronize the class-wise training losses.
We proposed a new training loss to reduce the variation of class-wise training losses, which can keep the entire training loss while reducing the calibration errors.
Furthermore, the pre-trained model can be used as an initial model and the extra computation for model calibration is reduced when our method is applied to post-hoc calibration methods.
The numerous post-hoc calibration approaches we use to test the proposed architecture generally increase calibration performance while maintaining accuracy.
The proposed algorithm showed the state-of-the-art performance in terms of calibration errors, even with various types of benchmark datasets and model architectures.
We are planning to apply the proposed method for the various applications the benefited by the well-calibrated predictions in our future work.

\section{Limitation}
Our proposed approach requires the supplementary parameters and a sigmoidal module, thereby resulting in an increase in computational complexity and an additional training process. Furthermore, to compute the class-wise loss for individual classes, we only utilize outputs that contain prediction information for each class. Consequently, it is important to note that the proposed method’s applicability may be restricted if the pre-trained model estimates an output in a form other than logits.

\section*{Acknowledgement} 
\footnotesize
This work was partly supported Institute of Information $\&$ Communications Technology Planning $\&$ Evaluation (IITP) grant funded by the Korea government (MSIT) (No. 2021-0-02067, Next Generation AI for Multi-purpose Video Search$;$ 2021-0-01341, Artificial Intelligence Graduate School Program(Chung-Ang University)), and a grant (22193MFDS471) from the Ministry of Food and Drug Safety in 2023.
\normalsize

\bibliographystyle{icml2023}
\bibliography{paper}

% APPENDIX
\newpage
\appendix
\onecolumn

\section{Loss Normalization}\label{Appendix:A}
\begin{table*}[t]
\caption{\label{tab: norm}Comparison result of Loss Normalization}
\label{norm}
\resizebox{\textwidth}{!}{%
\begin{tabular}{ c | c || c c c c | c c c c | c c }
\hline
\multirow{2}{*}{Calibration} & \multirow{2}{*}{Normalization} & \multicolumn{4}{c|}{CIFAR10} & \multicolumn{4}{c|}{CIFAR100} & \multicolumn{2}{c}{ImageNet} \\

& & DenseNet40 & WideResNet32 & ResNet110 & ResNet110SD & DenseNet40 & WideResNet32 & ResNet110 & ResNet110SD & ResNet152 & DenseNet161  \\
\hline\hline
\multirow{4}{*}{TS} 
& base  & 0.96 & 0.82 & 1.14 & 0.54 & 0.97 & 1.49 & 2.47 & 1.27 & 2.07 & 1.96 \\
\cline{2-12}
& ND    & \textbf{0.71} & \textbf{0.78} & \textbf{0.85} & \textbf{0.49} 
        & \textbf{0.81} & \textbf{1.23} & \textbf{1.84} & \textbf{1.18} 
        & \textbf{2.01} & \textbf{1.93} \\
& CM    & 0.86 & \textbf{0.78} & 1.03 & 0.53 & 0.94 & 1.38 & 2.41 & 1.24 & 2.07 & 1.95 \\
& MM    & 0.91 & 0.79 & 1.03 & 0.54 & 0.94 & 1.40 & 2.44 & 1.27 & 2.07 & 1.98 \\
\hline
\multirow{3}{*}{ETS} 
& base  & 0.97 & 0.81 & 1.14 & 0.54 & 0.97 & 1.49 & 2.47 & 1.28 & 2.06 & 1.96 \\
\cline{2-12}
& ND    & \textbf{0.71} & \textbf{0.76} & \textbf{0.86} & \textbf{0.47} 
        & 0.85 & \textbf{1.15} & \textbf{1.85} & \textbf{1.02} 
        & \textbf{2.02} & 1.96 \\
& CM    & 0.86 & 0.78 & 1.02 & 0.55 & \textbf{0.79} & 1.28 & 2.28 & 1.22 & 2.03 & \textbf{1.93} \\
& MM    & 0.87 & 0.77 & 1.02 & 0.53 & \textbf{0.79} & 1.35 & 2.40 & 1.22 & 2.04 & 1.95 \\
\hline
\multirow{3}{*}{PTS} 
& base  & 0.95 & 0.78 & 1.13 & 0.56 & 0.90 & 1.47 & 2.38 & 1.21 & 2.08 & 1.94 \\
\cline{2-12}
& ND    & \textbf{0.71} & \textbf{0.77} & \textbf{0.82} & \textbf{0.47} 
        & \textbf{0.80} & \textbf{1.18} & \textbf{1.88} & \textbf{1.05} 
        & \textbf{2.03} & \textbf{1.92} \\
& CM    & 0.85 & 0.79 & 1.02 & 0.55 & 0.80 & 1.35 & 2.44 & 1.18 & 2.04 & 1.96 \\
& MM    & 0.86 & 0.78 & 1.03 & 0.55 & 0.84 & 1.36 & 2.43 & 1.19 & 2.04 & 1.95 \\
\hline
\multirow{3}{*}{CTS} 
& base  & 1.01 & 0.84 & 1.31 & 0.75 & 0.99 & 1.61 & 2.50 & 1.15 & 2.38 & 2.24 \\
\cline{2-12}
& ND    & \textbf{0.74} & \textbf{0.44} & \textbf{1.04} & 0.61 
        & \textbf{0.74} & \textbf{1.42} & \textbf{1.99} & \textbf{0.97} 
        & \textbf{1.68} & \textbf{1.60} \\
& CM    & 0.85 & 0.74 & 1.11 & \textbf{0.57} & 1.03 & 1.50 & 2.24 & 1.20 & 1.96 & 1.91 \\
& MM    & 1.07 & 0.77 & 1.13 & 0.67 & 1.09 & 1.56 & 2.10 & 1.26 & 2.16 & 1.91 \\
\hline
\end{tabular}}
\end{table*}

% Through the experiment in Section.~\ref{section:4.1} of the main paper, we validated the balance between class-wise losses and the model's confidence. 
% Based on this, we assign a higher scale to classes with high uncertainty in their loss and a lower scale to classes with low uncertainty in their loss, and train the overall loss accordingly. 
% The scale is determined based on the relative size of the class-wise loss, and normalization is needed to calculate this relative size.
% Thus, we require normalized class-wise losses as input factors for the Class-wise Loss Scaling Estimator in order to balance the class-wise loss. 
Our idea is to achieve a balance of class-wise training losses. However, when the class-wise training losses are too low to be compared directly, balancing the class-wise training losses becomes a challenge. To address this issue, we use normalization methods.
We conducted experiments on various combinations of benchmarks using our approaches with three loss normalization methods: Min-Max normalization (MM), Centered Min-max normalization (CM), and Normal Distribution normalization (ND). 
Based on the class-wise training loss $\mathcal{L}_{C_i}$ defined in Eq.~\ref{eq:7}, we describe the following equations for the normalization methods.

\textbf{Normal Distribution normalization}\\
ND is also called Gaussian distribution normalization or standardization.
ND is calculated by subtracting the mean of the distribution from the value and dividing it by the standard deviation of the distribution, which is defined as follows:
\begin{equation}
ND(\mathcal{L}_{C_i}) = \frac{\mathcal{L}_{C_i} - \mathcal{L}_{C_{mean}}}{\mathcal{L}_{C_{std}}},
\end{equation}
where $\mathcal{L}_{C_{mean}}$ and $\mathcal{L}_{C_{std}}$ are mean and standard deviation of class-wise training loss, respectively.
ND does not have fixed min-max scales but provides robustness against outliers.

\textbf{Min-Max normalization}\\
MM is calculated by subtracting the minimum value of the distribution and dividing it by the gap between maximum and minimum of the distribution, which is defined as follows:
\begin{equation}
MM({\mathcal{L}}_{C_i}) = \frac{\mathcal{L}_{C_i} - \mathcal{L}_{C_{min}}}{\mathcal{L}_{C_{max}}-\mathcal{L}_{C_{min}}},
\end{equation}
where $\mathcal{L}_{C_{min}}$ and $\mathcal{L}_{C_{max}}$ are minimum and maximum values of class-wise training loss, respectively.
In contrast to ND, MM always has the same min-max scale ranging from zero to one but is weak against outliers.

\textbf{Centered Min-max normalization}\\
CM is a modified form of MM that alters the calculation by subtracting the mean instead of the minimum value, which is defined as follows:
\begin{equation}
CM(\mathcal{L}_{C_i}) = \frac{\mathcal{L}_{C_i} - \mathcal{L}_{C_{mean}}}{\mathcal{L}_{C_{max}}-\mathcal{L}_{C_{min}}}.
\end{equation}
CM maintains the same scale with MM, while showing the variations in the mean of the distribution. However, CM suffers from the weak robustness in handling outliers yet.

\textbf{Quantitative results}\\ 
In Table~\ref{tab: norm}, it can be observed that the empirical performance of ND is generally better than that of MM, except for three cases: ETS-CIFAR100-DenseNet40, ETS-ImageNet-DenseNet161, and CTS-CIFAR10-ResNet110SD.
We observed that poor performance in these cases was due to outliers, which had a negative impact on the scores of CM and MM. 
In contrast, ND demonstrated robustness against outliers, resulting in its superior performance. 
Therefore, ND outperforms CM and MM, making it a suitable choice for our approach.

\section{Comparison Methods}\label{Appendix:B}
Given a pre-trained network $\phi_o:=\mathbf{sm}\circ\mathbf{g}$ (where $\mathbf{g}$ is the classifier and $\mathbf{sm}$ is the softmax function) and datasets $\mathcal{D}=\{(\mathbf{x}^i,y^i)\}_{i=1}^N$, we define the output logit from the network by $\mathbf{z}^i=\mathbf{g}(\mathbf{x}^i)$. Now we define various methods based on \textit{Temperature Scaling} to calibrate given output logit. All cases except for one case preserve the model accuracy.\\

\noindent\textbf{Temperature Scaling} (TS)~\citelatex{TS} is the simplest method in temperature based methods. TS calibration function just divides logits into the same Temperature parameters as follows:
\begin{equation}
\mathbf{TS}(\mathbf{z}):= \frac{\mathbf{z}}{T}=\{\frac{\mathtt{z}_1}{T},...,\frac{\mathtt{z}_i}{T}\,...,\frac{\mathtt{z}_n}{T}\},
\end{equation}
where $i$ is the $i$-index vector corresponding with the class. 
Temperature is a simple tool to control the uncertainty of the model prediction. 
As T is close to infinity, all vectors from logit become equal, and as T is close to zero, the most prominent vector from logit becomes near infinity. 
TS prevents the change order between logits due to monotonic function. 
Thus, we can preserve the model's accuracy.
Furthermore, training latency is the fastest method because TS  train only one parameter.\\

\noindent\textbf{Ensemble Temperature Scaling} (ETS)~\citelatex{ETS} 
uses the ensemble model of TS. Increasing the number of models makes the prediction probability well-calibrated, but the cost also increases.
Thus, due to efficiency, Zhang proposed simplifying ETS into three terms follows:

\begin{equation}
\mathbf{ETS}(\mathbf{z}):= w_1\mathbf{TS}(\mathbf{z}) + w_2\mathbf{z} + \frac{w_3}{L} ,
\end{equation}
where $L$ is the number of classes. ETS calibration function consists of three-term, TS function, logits, and constant with weight parameters. Training of ETS separates two stages, temperature scaling with parameter $T$ and ensemble scaling with parameters $w_1$, $w_2$, and $w_3$.
ETS also preserves the model's accuracy due to consisting of monotonic functions.\\

\noindent\textbf{Parameterized Temperature Scaling}(PTS)~\citelatex{PTS} estimates Temperature parameters through a multi neural layer. 
PTS calibration function is defined as follows:
\begin{equation}
\mathbf{PTS}(\mathbf{z}):= \frac{\mathbf{z}}{h_{\theta_{ln}}(\mathbf{z_s})},
\end{equation}
where $\mathbf{z}_s$ is the sorted and selected top $s$ logits from $\mathbf{z}$, $h_{\theta_{ln}}$ function is a multi-neural layer in which the input size is $s$ and the output size is 1, and the multi-neural layer consists of $n$ nodes and $l$ fully connected layers.
The notion that imposes the same scale to logit is equal to the TS method. The difference is that PTS imposes an estimated T per sample by considering the correlation between top k vectors, while TS imposes T on all samples equally. Diverse T gives more expression to logit, so make well-calibrated logit than TS.\\

\noindent\textbf{Class-based Temperature Scaling} (CTS)~\citelatex{CTS} CTS is the extended form of TS by increasing the Temperature parameter to the same as the number of classes.
CTS calibration function is defined as follow:
\begin{equation}
\mathbf{CTS}(\mathbf{z}):= \{\frac{\mathtt{z}_1}{T_1},...,\frac{\mathtt{z}_i}{T_i}\,...,\frac{\mathtt{z}_n}{T_n}\},
\end{equation}

where $T_i$ is the class-based temperature for $i$-th class, which is a trainable parameters.
While other methods preserve the pre-trained model's accuracy, CTS cannot keep the accuracy for pre-trained prediction probabilities.
In some cases, however, CTS improves both model's accuracy and calibration due to no constraint.
\newpage

\begin{table*}[h]
\caption{\label{tab:detail info}Hyperparameters Details}
\label{ablation}
\resizebox{\textwidth}{!}{%
\begin{tabular}{c | c | c | c || c | c | c | c | c }
\hline
\multicolumn{4}{c||}{Information} & \multicolumn{5}{c}{Hyperparameter}\\
\hline
Datasets    & Architecture  & Accuracy   & Expected calibration error & Learning rate & Weight decay & Optimizer & Batch size & Epoch\\
\hline
\multirow{4}{*}{Cifar10-LT}
            & densenet40    & 72.11 & 21.08 & 0.01  & 1-e4  & SGD & 128 & 200\\ 
\cline{2-9} & WideResnet28  & 85.38 & 9.11 & 0.1   & 5-e4  & SGD & 128 & 200\\
\cline{2-9} & resnet110     & 75.97 & 19.14 & 0.1   & 1-e4  & SGD & 128 & 200\\
\cline{2-9} & resnet110sd   & 64.90 & 14.43 & 0.04  & 1-e4  & SGD & 128 & 200\\
\hline
\multirow{4}{*}{Cifar100-LT}
            & densenet40    & 37.97 & 38.4 & 0.01  & 1-e4  & SGD & 128 & 200\\
\cline{2-9} & WideResnet28  & 53.57  & 9.11 & 0.1   & 5-e4  & SGD & 128 & 200\\
\cline{2-9} & resnet110     & 38.57 & 41.98 & 0.1   & 1-e4  & SGD & 128 & 200\\
\cline{2-9} & resnet110sd   & 34.87 & 30.32 & 0.05  & 1-e3  & SGD & 128 & 200\\
\hline
\end{tabular}}
\end{table*}

\section{Imbalance Datasets}\label{Appendix:C}
Because our method considers the relation between class-wise loss, our method should show us a positive effect when unbalanced datasets exist about class.

\textbf{Long-Tailed Dataset}

We make a long-tailed dataset from CIFAR10 and CIFAR100 as follows~\citelatex{tang2020longtailed}.
Before the data split, we get each number of samples for each class defined as follows:

\begin{equation}
\#D_i=\#D_0 \times \rho ^{i/(c-1)},
\end{equation}

where $D_i$ is a dataset with label $i\in\{0,...,c-1\}$, $\#D_i$ is the number of samples for $D_i$, $\rho$ is the ratio to decide the minimum number of samples, and $c$ is the number of classes. Then, we bring data depending on the defined number of samples.
In CIFAR10, we bring 20431 samples from training data 50000 and separate them into 10215 training datasets and 10216 validation datasets.
We also bring 19573 samples from training data 50000 in CIFAR100 and separate them into 9786 training datasets and 9787 validation datasets.
The training dataset is used to learn the classifier, and the validation dataset is used to learn the calibration networks.

\textbf{Classifier Training for LT-dataset}\\
We train classifiers, which are densenet40~\citelatex{huang2017densely}, WideResNet28~\citelatex{BMVC2016_87}, ResNet110~\citelatex{7780459}, and ResNet110SD~\citelatex{10.1007/978-3-319-46493-0_39}, with obtained dataset above.
In the learning process, we use the hyperparameters referred to in Table.~\ref{tab:detail info}.
Next, we obtain the logit of the validation dataset and the test dataset through the learned classifier.
In the post-hoc calibration problem, the logit of the validation data and the logit of the test data are used for training and testing, respectively.

\bibliographystylelatex{icml2023}
\bibliographylatex{paper}

\end{document}